%% file: acl_latex.tex
\pdfoutput=1

\documentclass[11pt]{article}

\usepackage[final]{acl}

\usepackage{times}
\usepackage{latexsym}

\usepackage[T1]{fontenc}

\usepackage[utf8]{inputenc}

\usepackage{microtype}

\usepackage{inconsolata}

\usepackage{array,multirow}
\usepackage{graphicx}
\usepackage{bm}
\usepackage{amsmath}
\usepackage{amssymb}
\usepackage{amsfonts}
\usepackage{url}
\usepackage{algorithm}
\usepackage[noend]{algpseudocode}
\usepackage {arydshln}
\usepackage{comment}
\usepackage{CJKutf8}
\renewcommand{\u}{\underline}
\renewcommand{\b}{\textbf}

\title{Tokenization Preference for Human and Machine Learning Model:\\ An Annotation Study}

\author{
    Tatsuya Hiraoka \quad\quad Tomoya Iwakura \\
    Fujitsu Limited \\ 
    \texttt{\{hiraoka.tatsuya, iwakura.tomoya\}@fujitsu.com}
}

\begin{document}
\maketitle
\begin{abstract}
Is preferred tokenization for humans also preferred for machine-learning (ML) models?
This study examines the relations between \textit{preferred tokenization} for humans (appropriateness and readability) and one for ML models (performance on an NLP task).
The question texts of the Japanese commonsense question-answering dataset are tokenized with six different tokenizers, and the performances of human annotators and ML models were compared.
Furthermore, we analyze relations among performance of answers by human and ML model, the appropriateness of tokenization for human, and response time to questions by human.
This study provides a quantitative investigation result that shows that preferred tokenizations for humans and ML models are not necessarily always the same.
The result also implies that existing methods using language models for tokenization could be a good compromise both for human and ML models.
\end{abstract}

\input{latex/1_introduction}
\input{latex/2_relatedwork}
\input{latex/3_experimentalsettings}

\input{latex/4_result}
\input{latex/5_conclusion}

\clearpage
\section*{Limitation}
The scope of this study is limited to Japanese QA dataset for the reasons explained in \S \ref{sec:dataset}.
There is room for research in other languages and tasks (e.g., Chinese sentiment classification).
The QA architecture is limited to basic methods (i.e., BoW and BiLSTM). 
Further, this paper focuses on six tokenization methods (i.e., MeCab, Unigram, BPE, MaxMatch, OpTok, and Random).
We think these tokenization methods are well known and widely used for current NLP fields.

\section*{Ethics Statement}
For the further discussion with NLP researchers, we will open the corrected annotation as publicly available resource.
For the annotation, we recruited workers via a crowdsourcing service.
The purpose of annotation was told before work.
The payment was determined depending on the wage standard in the authors' country.

\section*{Acknowledgement}
This work was supported by JST, ACT-X Grant Number JPMJAX21AM, Japan.

\bibliography{custom,anthology}




\end{document}

%% file: latex/1_introduction.tex
\section{Introduction}

Tokenization is an fundamental preprocessing step that has a considerable effect on the performance of NLP systems such as text classification models.
The existing literature concludes that we can obtain performance improvement by selecting an appropriate tokenization method for the downstream task or model~\cite{hiraoka2019stochastic,bostrom2020byte,shin2020biomegatron}.
Recent work introduces methods that automatically select appropriate tokenizations for improving the performance of machine-learning (ML) models~\cite{salesky2020optimizing,xuanli2020dynamic,hiraoka2020optimizing,hiraoka2021joint}.

\begin{figure}[t]
\centering
\includegraphics[width=7.5cm]{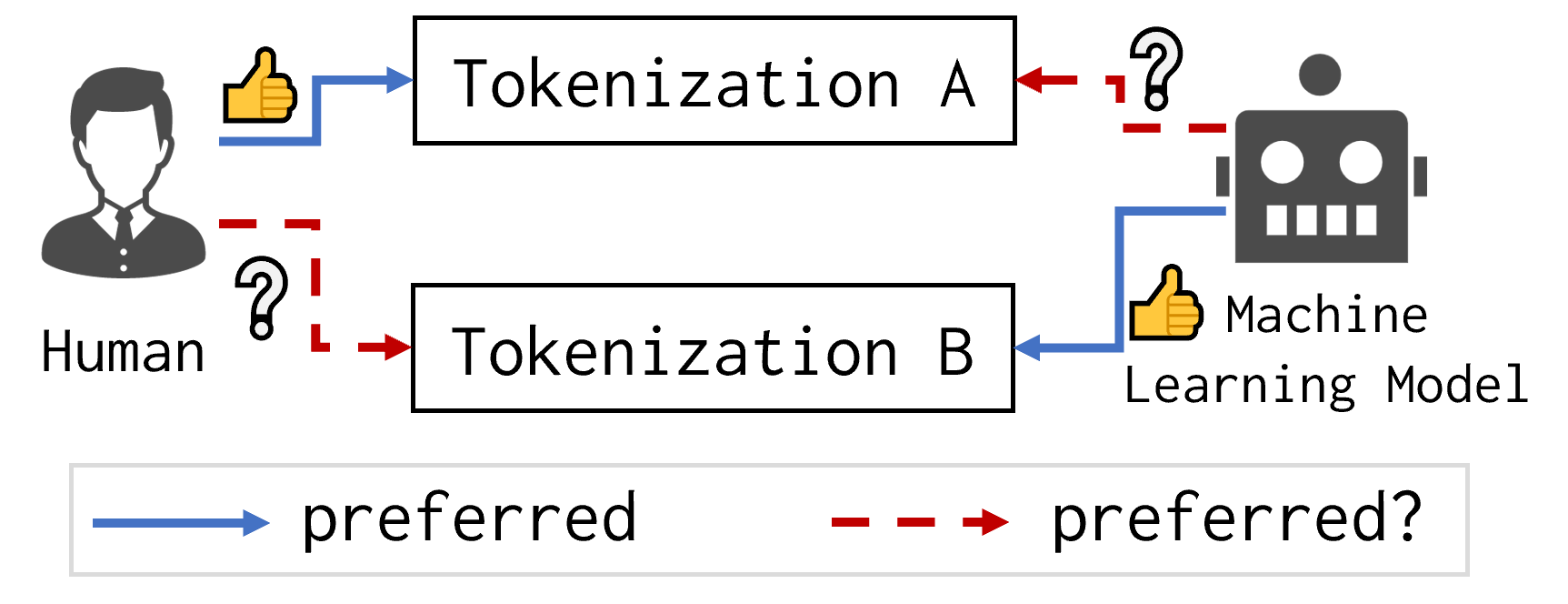}
\caption{
    We investigate whether a preferred tokenization for ML model might not always be the same as one for humans and vice versa.
}
\label{fgr:outline}
\end{figure}

Although previous work attempts to find appropriate tokenization for the ML model, it does not address or discuss whether such tokenization is preferred for humans (Figure \ref{fgr:outline}), especially from the quantitative viewpoint.
Some tokenization-related studies provide the statistical features of tokenization that improve the ML performance ~\cite{provilkov2019bpe,gowda2020finding,hiraoka2022maxmatch,zouhar2023tokenization}.
A study on optimizing tokenization provides small qualitative observation that the obtained tokenization examples appear inappropriate for readability~\cite{hiraoka2021joint}.
However, these works do not sufficiently discuss the preference of tokenization for humans and the use of human knowledge for further improvement.
We think that the investigation of tokenization preference for humans and ML models will provide insights into construction of tokenizers yielding preferred tokenization both for human and machine, also from the viewpoint of human-centered AI and human alignment.

This study provides a quantitative analysis of the relation between preferred tokenization for humans and ML models.
Focusing on the Japanese commonsense question–answering (JCommonsenseQA) in JGLUE~\cite{kurihara2022jglue}, we compared six tokenization methods (Table \ref{tbl:tokenizations}) from the viewpoint of humans and ML performance.
As for the performance of ML models, we assessed accuracy of models on the QA dataset with different tokenizatoins.
As for the performance of human, we collected the following three annotations with different tokenizations:
1) the appropriateness of the tokenization results of the question text by ranking, 
2) the accuracy of human response on the QA dataset, 
and 3) the response time for answering to each question.

Our contribution is to provide a quantitative analysis method and its results for further discussion on the tokenization preference for humans and machines.
The investigation result shows that preferred tokenization for humans is not necessarily equivalent to preferred tokenization for ML models and vice versa. 
The investigation result also shows that existing work language model-based tokenization and the method optimizing tokenization without referencing human knowledge could be an appropriate setting on this research task setting.

\input{tables/tokenizations.tex}

%% file: tables/tokenizations.tex
\begin{table}[t]
\centering
\small
\begin{tabular}{p{0.95cm}p{5.84cm}}
\hline
\textbf{Method} & \textbf{Tokenization} \\\hline
MeCab  &	\begin{CJK}{UTF8}{ipxm}もの\_を\_入れる\_容器\_を\_なんと\_言う\_?\end{CJK}        \\
Unigram &	\begin{CJK}{UTF8}{ipxm}もの\_を入れる\_容器\_を\_なん\_と言う\_? \end{CJK}       \\
BPE &	\begin{CJK}{UTF8}{ipxm}ものを\_入れる\_容器\_を\_なん\_と言う\_? \end{CJK}     \\
MaxMatch &	\begin{CJK}{UTF8}{ipxm}ものを\_入れる\_容器\_をな\_んと\_言う\_? \end{CJK}   \\
OpTok &	\begin{CJK}{UTF8}{ipxm}もの\_を\_入れる\_容器\_を\_なん\_と言う\_?    \end{CJK}   \\
Random &	\begin{CJK}{UTF8}{ipxm}も\_のを入\_れ\_る容\_器\_を\_なんと言\_う\_?    \end{CJK}   \\
\hline
\end{tabular}
\caption{
    Tokenization examples.
    ``\_'' indicates the token boundaries.
    The text means ``What do you call a container that holds things?''.
    \label{tbl:tokenizations}
}
\end{table}

%% file: latex/2_relatedwork.tex
\section{Related Work}
Word segmentation is regarded in the NLP field as a task to reproduce hand-crafted word boundaries for unsegmented languages such as Chinese~\cite{sproat-etal-1996-stochastic,xue-2003-chinese,peng-etal-2004-chinese,goldwater2009bayesian,mochihashi2009bayesian,zheng-etal-2013-deep}.
In the neural era, word segmentation (or tokenization) is considered as a preprocessing step for neural networks~\cite{sennrich2016neural,kudo2018sentencepiece,song2020linear}, which is not limited to hand-crafted word boundaries but seeking appropriate tokenization specialized to ML models.
Recent studies propose methods to find tokenizations for improving the downstream tasks~\cite{xuanli2020dynamic,hiraoka2020optimizing,hiraoka2021joint}.
We examined the relation between tokenization preference for humans and ML models, which is not focused on by the existing research.

In the linguistic literature, \textit{word} is defined as an independent, inseparable, meaningful, and smallest unit~\cite{marchand1969categories}.
A word boundary is discussed from phonological word and grammatical word~\cite{dixon2003word} for languages that do not have orthographic word boundaries indicated with whitespaces.
This study provides the first-step investigation in bridging the discussion of tokenization in NLP and word boundary in linguistics by analyzing the preference of tokenization for humans and ML models.

Recent work~\cite{beinborn-pinter-2023-analyzing} provides an analysis of response times and accuracy of a lexical decision task by human, which uses the similar method to our work.
Our work is different from their work in that we focus on the sentence-level tokenization in Japanese, which is much sensitive against tokenization difference because it does not use whitespaces indicating word boundaries

%% file: latex/3_experimentalsettings.tex
\input{tables/tokenizationInfo}

\section{Experimental Settings}
\subsection{Dataset}
\label{sec:dataset}
In this research, JCommonsenseQA in JGLUE~\cite{kurihara2022jglue} was used to investigate the preference of tokenization.
This is a dataset for a question–answering task of Japanese commonsense with one question text and five word choices.
The training split includes 8,939 questions and the validation split includes 1,119 questions (10,058 questions in total).
The original data splits were used in this experiment.
The test split was unavailable at the time of the experiment.

We chose this dataset because unlike English, Japanese texts are written without whitespaces to indicate word boundaries.
Therefore, the Japanese text can be tokenized in various ways with the differences in tokenization having a greater effect on the preference for human and ML model.
Moreover, unlike other Asian languages such as Chinese and Korean, most Japanese text includes several types of characters that have different information granularity (e.g., \textit{hiragana}, \textit{katakana} and \textit{kanji}\footnote{\textit{Hiragana} and \textit{katakana} are phonograms that do not have any meaning by itself (e.g., ``\begin{CJK}{UTF8}{ipxm}も\end{CJK}'' in Table \ref{tbl:tokenizations} pronounced as ``mo''). Conversely, \textit{kanji} is an ideograph that has meaning (e.g., ``\begin{CJK}{UTF8}{ipxm}入\end{CJK}'' pronounced as ``i'' means ``put-in'' by itself).}).
We consider that these characteristics of Japanese would bring a clear difference of the tokenization effect on preference.

We selected the commonsense QA task because annotators can answer the questions without any special knowledge.
Therefore, we can easily compare human and ML model performances.
The task is neither too easy for machines nor too difficult for humans.
This keeps the experiment from the external effects such as ML models solving the task only from the information of a few tokens (e.g., emotion classification can be solved from some polarity tokens) and human annotators cannot solve the task owing to the lack of special knowledge.

In addition to the QA dataset, we used a Japanese Wikipedia dump\footnote{\url{https://dumps.wikimedia.org/jawiki/20220820/}} to train several unsupervised tokenizers and to pretrain the module to calculate token embeddings from characters mentioned in the following sections (\S \ref{sec:compared_tokenizations} and \S \ref{sec:qa_systems}).

\subsection{Compared Tokenizations}
\label{sec:compared_tokenizations}
We compared six tokenization methods: dictionary-based tokenization (\textbf{MeCab}), unsupervised tokenization (\textbf{Unigram}, \textbf{BPE}, \textbf{MaxMatch}), a method optimizing tokenization to a ML model (\textbf{OpTok}), and random tokenization (\textbf{Random}).
We provide their information as the following.

\noindent\textbf{MeCab:}
We used MeCab~\cite{kudo2006mecab} for tokenization using a hand-crafted word dictionary.
For the dictionary, we used UniDic\footnote{\url{https://clrd.ninjal.ac.jp/unidic/about_unidic.html}} with which we obtain tokenization including many shorter tokens compared to other dictionaries (e.g., IPAdic\footnote{\url{https://ja.osdn.net/projects/ipadic/}} and mecab-ipadic-NEologd~\cite{sato2015mecabipadicneologd}).
We expect that this tokenization should be the most preferable (appropriate) for humans among the compared methods because it uses hand-crafted dictionary.

\noindent\textbf{Unigram:}
We used SentencePiece~\cite{kudo2018sentencepiece} with ``Unigram-mode''.
The tokenizer is trained to increase the likelihood of tokenization with the unigram language model.

\noindent\textbf{BPE:}
Byte-pair encoding (BPE) recursively merges tokens to make tokenization.
We used the ``BPE-mode'' of SentencePiece~\cite{kudo2018sentencepiece} for BPE tokenization.

\noindent\textbf{MaxMatch:}
A maximum matching-based tokenizer ~\cite{song2020linear} greedily tokenizes the input text with a vocabulary from the beginning to the end.
We used BertTokenizer known as WordPiece for this tokenization method.

\noindent\textbf{OpTok:}
OpTok~\cite{hiraoka2021joint} is a method for seeking appropriate tokenization for a ML model.
We used this method to the already trained QA model to obtain the optimized tokenization.
Because OpTok is based on the Unigram-based tokenizer, its vocabulary is completely the same to that of Unigram but tokenization is different.

\noindent\textbf{Random:}
We tokenize an input text into tokens from the begging to the end by randomly sampling the length of the token.
We counted the frequency of token lengths in the MeCab-tokenized Japanese Wikipedia data, and then used the token length distribution for the random sampling of length in the random tokenization.
We expect a randomly tokenized sequence to be the least preferable for both humans and ML models.

We set the number of tokens in the vocabulary to 64,000 for the unsupervised methods, \footnote{
In order to obtain similar total number of tokens on the QA corpus by each tokenization method, we selected the vocabulary size because the size leads to  almost same total number of tokens of MeCab on the corpus.}.
Then we built the tokenization models with the Japanese Wikipedia.
We did not use the training split of JCommonsenseQA to build the tokenization model because recent NLP systems usually utilize a tokenizer that is trained on an outside large corpus (e.g., the tokenizer bundled with large language models like $\mathrm{BERT}_\mathrm{base}$~\cite{devlin2018bert}).
Table \ref{tbl:tokenizations} shows the tokenization examples for each method introduced herein.

Table \ref{tbl:tokenizationInfo} summarizes the statistics of tokenization obtained from each method.
The top part of the table named ``Size of Vocabulary'' shows the number of token types that appear in the training, validation, and entire corpus of JCommonsenseQA, in addition to the size of vocabulary of unsupervised methods.
The middle part of the table named ``Tokenization Length'' reports the average number of tokens comprising question texts tokenized by each method.
The bottom of the table named ``Entropy'' shows the entropy of tokenization that is measured by counting tokens in each split\footnote{We constructed a new unigram language model by counting the frequency of tokens in each split and calculated the perplexity of the language model on the split.}.
Lower entropy indicates the peaky distribution of token frequency.

\section{Tokeniation Preference Annotation}
\label{sec:annotation}

\begin{figure}[t]
\centering
\includegraphics[width=7.5cm]{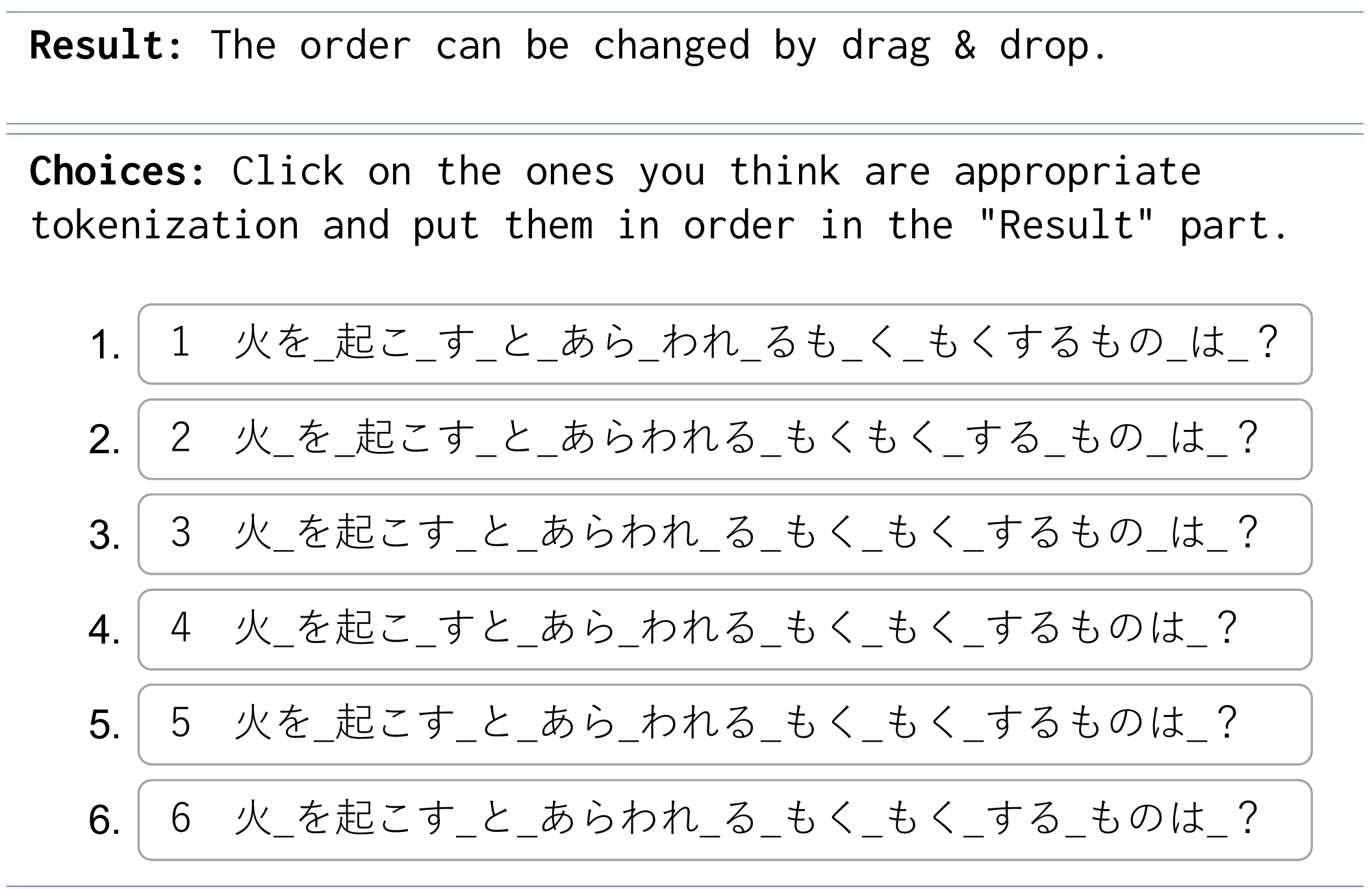}
\caption{
    Annotation tool for ranking readability.
    The Japanese text means ``What is that smoldering thing that appears when you build a fire?''.
    UI instruction texts are translated into English for the explanation.
}
\label{fgr:annotationScreenRanking}
\end{figure}

\begin{figure*}[t]
\centering
\includegraphics[width=15.0cm]{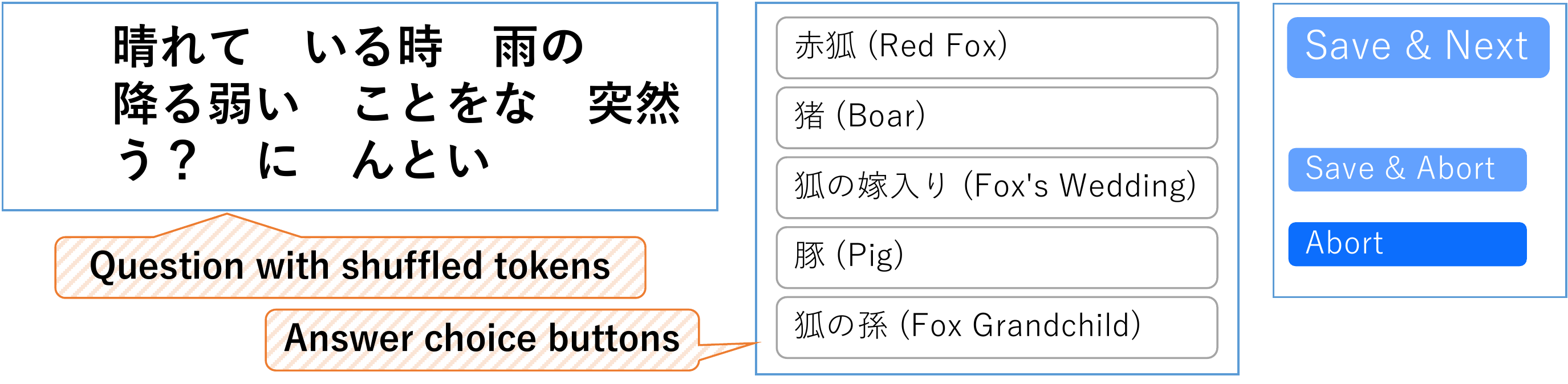}
\caption{
    Annotation tool for the bag-of-words question (left), answer candidates (center), and other UI buttons (right).
    The original order of the Japanese text is ``\begin{CJK}{UTF8}{ipxm}晴れて\_いる時\_に\_突然\_降る弱い\_雨の\_ことをな\_んとい\_う？\end{CJK}'' (random tokenization) meaning ``What do you call a weak rain that suddenly falls during a sunny day?''.
    The correct answer is ``\begin{CJK}{UTF8}{ipxm}狐の嫁入り\end{CJK}'' meaning ``Fox's Wedding''.
    The texts on the UI buttons are translated into English for the explanation.
}
\label{fgr:annotationScreen}
\end{figure*}

We collected information about tokenization preference for humans by annotation.
Herein, we conducted the annotation for the entire samples of a question–answering task, JCommonsenseQA (See \S \ref{sec:dataset}), from two perspectives: appropriateness ranking (\S \ref{sec:ranking_anno}) and the actual human response to the QA (\S \ref{sec:speed_anno}).
All annotations were conducted using a crowdsourcing service.
Further, we selected annotation clues that are not specialists in linguistics to prevent the result from being affected by the special knowledge of each annotator.

\subsection{Appropriateness Ranking}\label{sec:ranking_anno}
Ranking annotation of the tokenization appropriateness is the simplest way to collect differences of preference in tokenization.
The annotators were given a maximum of six different tokenizations of each question text in the dataset.
The annotators were then instructed to sort the tokenizations into the order of appropriateness.
Figure \ref{fgr:annotationScreenRanking} shows a screenshot of the annotation tool.
Notably, the tokenizations were shown in the shuffled order for each annotation sample.
We focused on only question texts of the QA dataset for the annotation  because all the answer choices are single words or very short phrases with few tokenization candidates (see choices in the center box of Figure \ref{fgr:annotationScreen}).
In this annotation, a single annotator is assigned to conduct the ranking of the entire dataset.
Besides, other two annotators are assigned for the validation split of the dataset.
In other words, the training split has a single annotation and the validation split has three annotations.

\subsection{Answering Bag-of-Words Questions}\label{sec:speed_anno}
The result of the annotation method in \S \ref{sec:ranking_anno} may be inconsistent because it has been difficult to define what is appropriate tokenization in this research.
In fact, the obtained annotation faced this problem (see \S \ref{sec:acc_vs_ranking}).
Moreover, the annotation results would strongly depend on the linguistic knowledge of the annotator even if they are not a linguistic specialist.
For example, if the annotator knows a discussion of morphology, they will assign the morphologically reasonable tokenization as the appropriate tokenization.
However, herein, we need to collect tokenizations that are preferred to solve the downstream task (i.e., QA) for humans to compare them with preferred tokenization for ML models.
Considering this point, the annotation collected in \S \ref{sec:ranking_anno} is not yet sufficient.

To accomplish this goal, we additionally collected the actual answers to the questions and the response time by annotators\footnote{We followed the existing work that collects annotation using texts and images~\cite{ma2022ontheeffectiveness}}.
Annotators were given a question and buttons with answer choices, and they are instructed to select one choice that is suitable to the given question.
We cannot collect the difference in response caused by tokenization if we simply show the tokenized question to the annotators like the ranking annotation because annotators easily understand the meaning regardless of the tokenization difference.
For example, when a text ``interesting'' is tokenized in two ways and shown in the original token order: ``inter\_est\_ing'' and ``int\_eres\_ting'', English native speakers can easily reproduce the original meaning.
Therefore, as shown in the left-box of Figure \ref{fgr:annotationScreen}, we shuffled the token order of the question text such as ``est\_ing\_inter'' and ``eres\_int\_ting''.
From the above, we hypothesized that if a text is appropriately tokenized, humans can reproduce the original meaning of the text even if the word order of the text is shuffled.

The setting of token shuffling can be seen as the bag-of-words setting of the QA task for human annotators because the information on the token order is removed.
This allows us to compare the results of human and ML methods in a fairer environment for the bag-of-words model.

In addition to the answers by annotators, we also measured the response time from the appearance of the annotation page to the pushing of the button of choices\footnote{The end point of response time is measured at the moment of the final push of a choice button, instead of pushing ``Save \& Next''. We ignored the pushes of the button before the final choice because annotators sometimes push some buttons without any meaning.}.
Based on the above hypothesis, we expect that the response time for the more appropriate tokenization will be shorter.

We were unable to monitor every action of the work because we used crowdsourcing for the annotation.
Therefore, herein, the annotation of response time that is longer than 30 s was removed~\footnote{Native Japanese speakers can answer most questions in this dataset around 5 s. Even for the question with a long text, it takes around 15 s (Figure \ref{fgr:boxplot}).} because annotators might leave the annotation screen untouched for these questions.

Six annotators worked on the annotation of the bag-of-words QA task.
To remove the effect of annotation caused by individual differences between annotators, we assigned all tokenization variations of each question to a single annotator.
In other words, a single annotator answers to the six different tokenization (i.e., MeCab, Unigram, BPE, MaxMatch, OpTok, and Random) for each question.
This causes a problem that the annotator can memorize the contents and answers to the questions because the annotator would annotate the same question up to six times with different tokenization results.
To alleviate this issue, we gave the same question to the annotator with some intervals (averaged 1591.8 questions).
All the data samples (10,058 questions) are answered by two different annotators.

\section{QA Architecture}
\label{sec:qa_systems}
The purpose of this study is to examine and compare the preference of tokenization between humans and ML models.
As the first step of this investigation, we use simpler architectures for the QA model.
Herein, we used two simple QA methods: (1) a bag-of-words-based method~\cite{boyd2012besting} and (2) a BiLSTM-based method~\cite{hochreiter1997long, graves2005framewise}.

Given a question comprising $N$ tokens $q = w_1, ..., w_N$, we calculate the probability $p(a|q)$ that a choice $a$ is the answer of $q$ as the following:
\begin{align}
\mathbf{v}_q &= g(f(\mathbf{v}_{w_1}, ..., \mathbf{v}_{w_{N}})), \\
p(a|q) &\sim \mathbf{v}_q^\top \mathbf{v}_a,
\end{align}
where $f(\cdot)$ is an encoder and $g(\cdot)$ is an MLP that transforms the output of the encoder into the size of token vectors.
$\mathbf{v}_{w_n}$ and $\mathbf{v}_a$ are token vectors correspond to $w_n$ and $a$, respectively.

We use the $f_{\mathrm{BoW}}$ and $f_{\mathrm{BiLSTM}}$ as $f(\cdot)$ above for the bag-of-words-based QA system and the BiLSTM-based QA system, respectively.
\begin{align}
f_{\mathrm{BoW}}&(\mathbf{v}_{w_1}, ..., \mathbf{v}_{w_{N}}) = \frac{\sum_{w \in s}{\mathbf{v}_{w}}}{N}, \\
f_{\mathrm{BiLSTM}}&(\mathbf{v}_{w_1}, ..., \mathbf{v}_{w_{N}}) \nonumber \\
    &= h(\mathrm{BiLSTM_{QA}}(\mathbf{v}_{w_1}, ..., \mathbf{v}_{w_{N}})), 
\end{align}
where $\mathrm{BiLSTM_{QA}}(\cdot)$ is a function that yields time step-wise output of a BiLSTM layer, and $h(\cdot)$ is the maximum pooling of the BiLSTM outputs.

\begin{figure}[t]
\centering
\includegraphics[width=7.5cm]{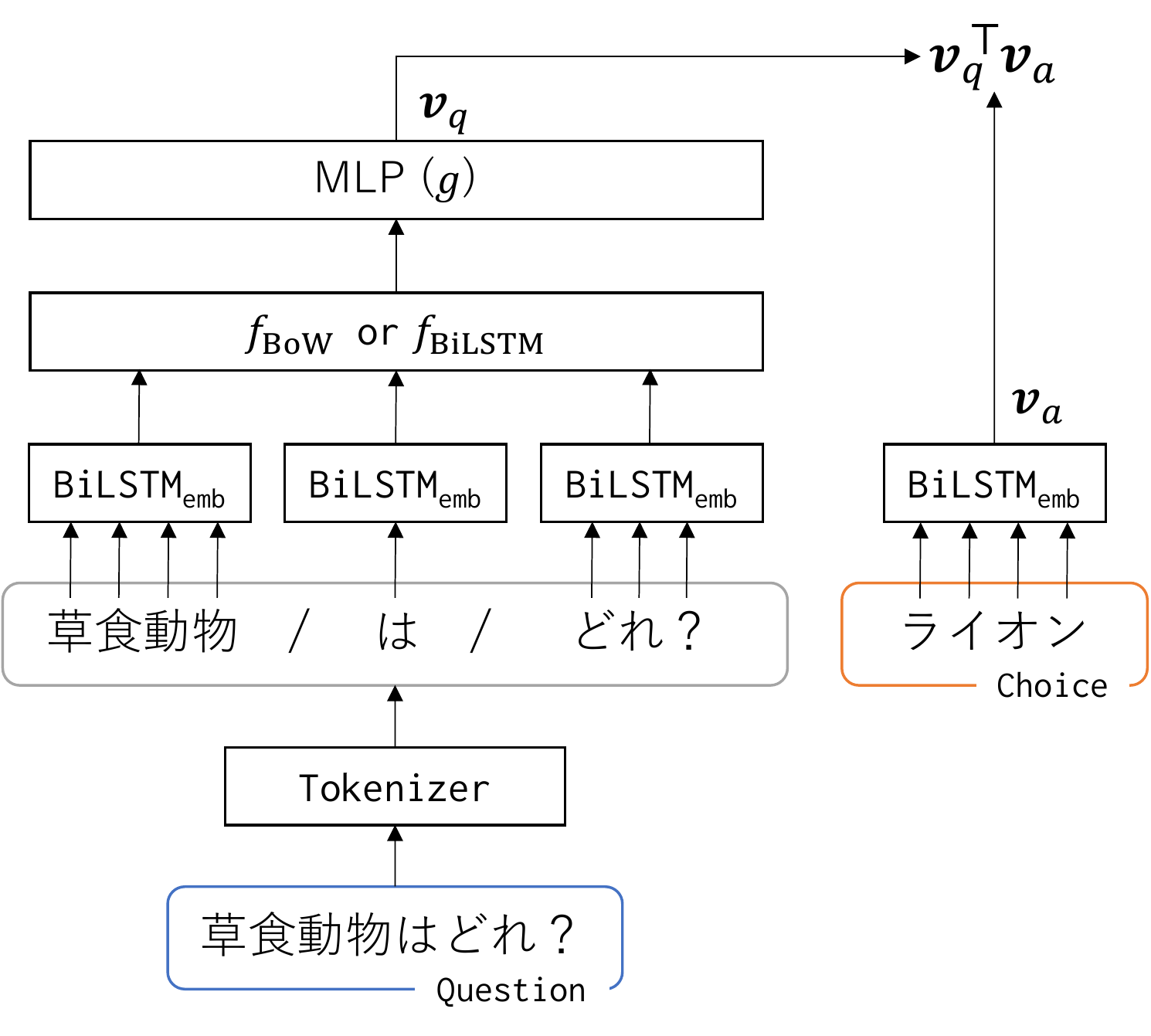}
\caption{
    Outline of the QA system using bag-of-words (BoW) or BiLSTM to calculate the score for the question ``\begin{CJK}{UTF8}{ipxm}草食動物はどれ？\end{CJK}'' (which animals are herbivores?) and the choice ``\begin{CJK}{UTF8}{ipxm}ライオン\end{CJK}'' (lion).}
\label{fgr:qa_system}
\end{figure}

We calculate the token vectors $\mathbf{v}_{w_n}$ and $\mathbf{v}_a$ from the character embeddings~\cite{lample2016neural,wang2017convolutional} because we need to exclude non-tokenization factors that might affect the performance.
Specifically, we use the same method to obtain token embeddings with the same character embeddings over different tokenizations.
The token vector $\mathbf{v}_w$ composed of $M$ characters $w = c_1, ..., c_M$ is calculated as follows:
\begin{align}
\mathbf{v}_{w} &= h(\mathrm{BiLSTM_{emb}}(\mathbf{v}_{c_1}, ..., \mathbf{v}_{c_M})),
\end{align}
where $\mathbf{v}_{c_m}$ is a word embedding corresponding to the $m$-th character $c_m$.
Similarly, we calculate the token vector $\mathbf{v}_a$ corresponding to the answer choice $a$.
Figure \ref{fgr:qa_system} outlines this calculation. 

The size of the embedding $\mathbf{v_c}$ was 256.
We set the size of hidden vectors of $\mathrm{BiLSTM_{emb}}$ and $\mathrm{BiLSTM_{QA}}$, and the size of $\mathbf{v}_w$  as 1,024.
We used a single-layered BiLSTM for both $\mathrm{BiLSTM_{emb}}$ and $\mathrm{BiLSTM_{QA}}$.
The encoders $f_{\mathrm{BoW}}$, $f_{\mathrm{BiLSTM}}$, and the MLP $g$ were trained on the training split.
We pretrained $\mathbf{v}_{c_m}$ and $\mathrm{BiLSTM_{emb}}$ on the entire data of the Japanese Wikipedia in the same manner as the character-level bidirectional modeling~\cite{peters2018deep}, and they were frozen in the training of the QA task.
We used cross-entropy loss and Adam~\cite{kingma2014adam} for the training of the QA task.
The QA models were trained in a maximum of 30 epochs, and we selected the best-performing model on the validation split.

%% file: tables/tokenizationInfo.tex
\begin{table*}[ht]
\centering
\small
\begin{tabular}{lrrrrrr}
\hline
                               & MeCab      & Unigram & BPE        & MaxMatch   & OpTok  & Random         \\ \hline
\textit{\textbf{Size of Vocabulary}}    &            &         &            &            &        &                \\
\# of Vocab in Tokenizer             & -          & 64,000  & 64,000     & 64,000     & 64,000 & -              \\
\# of Vocab in Train            & 8,658       & 11,172   & 11,534      & 12,090      & 10,709  & 17,473          \\
\# of Vocab in Valid            & 2,459       & 3,141    & 3,321       & 3,431       & 3,021   & 3,949           \\
\# of Vocab in Total            & 9,267      & 11,921  & 12,292     & 12,890     & 11,440 & 19,077         \\ \hline
\textit{\textbf{Tokenization Length}} &            &         &            &            &        &                \\
Train                          & 10.22      & 8.04    & 7.79       & \u{7.73} & 8.41   & \textbf{10.59} \\
Valid                          & 10.17      & 8.00    & 7.78       & \u{7.72} & 8.35   & \textbf{10.49}  \\
Total                          & 10.21       & 8.03    & 7.79       & \u{7.73} & 8.40   & \textbf{10.58}  \\ \hline
\textit{\textbf{Entropy}}               &            &         &            &            &        &                \\
Train                          & \u{8.27} & 9.68    & 10.18      & 10.23      & 9.35   & \textbf{10.39} \\
Valid                          & \u{7.90} & 9.06    & 9.51       & 9.54       & 8.78   & \textbf{9.75}  \\
Total                          & \u{8.29} & 9.71    & 10.20      & 10.25      & 9.38   & \textbf{10.43} \\ \hline
\end{tabular}
\caption{
    Statistics of each tokenization method.
    Size of vocabulary, averaged token length, and entropy are measured after tokenizing each data split with the tokenizers.
    Bold and underline indicate the highest and lowest value among each row. respectively.
    \label{tbl:tokenizationInfo}
}
\end{table*}


%% file: latex/4_result.tex
\input{tables/results}

\section{Result and Analysis}
\label{sec:results}
\subsection{Overview of Result}
Table \ref{tbl:result} summarizes the experimental results with QA models and annotators.
``Accuracy of Machine'' in the table shows the experimental results of QA models with bag-of-words (BoW) and BiLSTM on the validation split for three runs.
The machine-learninig model trained with unsupervised-based tokenization (i.e., Unigram, BPE, MaxMatch, and OpTok) showed higher accuracy compared to the models with MeCab and Random.

The second part named ``Appropriateness'' shows the ranking annotation result of the training, validation, and entire splits (\S \ref{sec:ranking_anno}).
We showed the averaged ranks of each method. 
Notably, some question texts are tokenized in the same way by several tokenization methods (e.g., Unigram and BPE yield the same tokenization for a text in some cases).
We assigned the same ranking score for such tokenization samples.
The best appropriateness was obtained with MeCab in most cases.

The third and fourth parts show the accuracy of human and the response time for the annotation, respectively (\S \ref{sec:speed_anno}).
The highest accuracy by annotators on the validation split was 97.5 with MeCab tokenization.
Considering the annotation environment, where tokens in the question texts are shown in the shuffled order, this performance is reasonable when compared with to the officially reported human accuracy: 98.6~\cite{kurihara2022jglue}.

We show the response time for all answers and the one only for correctly answered samples.
The result shows that the correctly answered samples took shorter response time compared to the entire samples.
This implies that understandable tokenization contributes to the shorter response time, and this supports our assumption.

From these results, we see that MeCab with a handcrafted dictionary is the most preferred for human as for accuracy, whereas unsupervised methods lead to the higher accuracy for the ML models.
Therefore, we conclude that preferred tokenization for humans is not necessarily equivalent to one for ML models and vice versa.
On the other hand, interestingly, we also observe that the response time for MeCab tokenization was not the shortest, which does not match the result of accuracy.
We further delve into the results in the following subsections.

\subsection{Machine Accuracy vs. Appropriateness}
\label{sec:acc_vs_ranking}
``Accuracy of Machine'' in Table \ref{tbl:result} shows that the performance of the BoW scores the best with BPE, and second with Unigram.
The BiLSTM-based model performs best with Unigram and MaxMatch.
The scores of both QA models for Random and MeCab tokenization have considerably decreased.

The result of annotation in ``Appropriateness'' shows that the question texts tokenized with MeCab are annotated as the most appropriate in most splits by each annotator.
Because MeCab is the tokenizer that uses the hand-crafted dictionary, the result is equivalent to what we expected as the hypothesis (\S \ref{sec:compared_tokenizations}).
Unigram and OpTok tend to be selected as second and third appropriate tokenizations.
This shows that tokenization methods with the language model are preferred for humans among other tokenization methods (i.e., BPE and MaxMatch).
This seems to be related to the existing report that tokenization yielded by a tokenizer based on a language model match the ones by humans even in the environment of unsupervised word segmentation~\cite{mochihashi2009bayesian}.

The performance of QA models with MeCab is the second lowest result for both BoW and BiLSTM, while MeCab is the most appropriate tokenization for humans by a large margin.
This result shows that preferred tokenization for the ML model differs from the most preferred one for humans.
With the exception of MeCab, Unigram is selected as the appropriate tokenization in many samples by humans and it also improves the ML performance.
This inplies that Unigram could be a good compromise between humans and machine preference of tokenization.

One remarkable point of the human annotation is that ``Appropriateness'' of MeCab on the validation split by Annotator1 was the most inappropriate among the methods excluding Random, in contrast to the overall tendency.
We discovered that this is caused by the annotation inconsistency.
The standard for what constitute appropriate tokenization was gradually shifted in the annotation, especially in the last part, which includes the validation samples.
This fact indicates the difficulty to define what is appropriate tokenization.
Facing to this problem, we additionally hired two other annotators as Annotator2 and 3, and collect ranking annotation for the validation split.
From the result by three annotators, we can finally say that MeCab could be the most appropriate tokenization for human.

\subsection{Accuracy of Human vs. Machine}
\label{sec:accH_accM}
The accuracy of humans reported in ``Accuracy of Human'' in Table \ref{tbl:result} does not show a considerable difference among tokenization methods.
However, the performance for Random is relatively lower than the other methods.
This tendency is similar to the performance of ML models shown in ``Accuracy of Machine''.
We think that the dataset is easy to solve even when the tokens of texts are shuffled.

Although we did not see the remarkable differences in human performance, MeCab showed the highest accuracy for both annotators in total.
This trend is different from the ML models whose scores with MeCab were lower than most tokenization methods, implying mismatching of tokenization preference between human and machine.


\subsection{Relation between Answer Time and Tokenization}
We observed small differences in the accuracy by human for different tokenizations (\S \ref{sec:accH_accM}).
We examine tokenization acceptability from the viewpoint of the response time~\cite{beinborn-pinter-2023-analyzing}.

``Response Time'' in Table \ref{tbl:result} shows that the response time for the question was short when the question text is tokenized with Unigram.
This result demonstrates that human annotators can quickly respond to the question that is tokenized with language model-based tokenizers.
Unigram-based method, OpTok, also contributes to the shorter response time.
OpTok tends to cut off function words like postposition particles compared to the tokenization of Unigram~\cite{hiraoka2021joint}.
This could make the shuffled tokenization easier to be reconstructed into the original meaning and contribute to a faster response. 

Random tokenization considerably increases the response time particularly long in all data splits.
This indicates that the random tokenization is not appropriate to understand the question because it is difficult for annotators to reconstruct the original meaning of question texts from the shuffled tokens.

\begin{figure}[t]
\centering
\includegraphics[width=7.5cm]{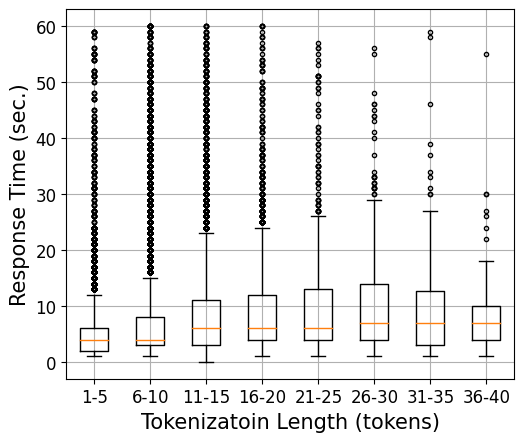}
\caption{
    The box plot for the relation between tokenization length and response time over all annotation results. We measured it for entire annotations mixing all tokenizers but excluded the sample whose response time is longer than 60 s.
}
\label{fgr:boxplot}
\end{figure}

Interestingly, the response time for MeCab tokenization is not the shortest in many cases.
This result could be owing to the feature of MeCab tokenization that produces longer sequences (``Tokenization Length'' in Table \ref{tbl:tokenizationInfo}).
When the question text is tokenized into many numbers of tokens and is shuffled, annotators need a much longer time to reconstruct the original token order.
The correlation coefficient between the response time and tokenization length was 0.22, indicating a weak positive correlation.
We visualized the relation between tokenization length and response time for entire annotations as the box plot in Figure \ref{fgr:boxplot}.

Moreover, there is another fact that the response speed for OpTok was the shortest for Annotator1 while the tokenization length of OpTok is relatively long.
Therefore, we consider that there are other reasons that affect the response time other than the tokenization length.
For example, the annotators might think deeply because they can understand the meaning of questions with MeCab, resulting in a lengthy response time.
The MeCab tokenization includes a much larger numbers of appropriate tokens, and annotators can consider many possibilities of choices, while they might select choices by their intuition without thinking anything for the inappropriately tokenized questions.

%% file: tables/results.tex
\begin{table*}[ht]
\centering
\small
\begin{tabular}{lrrrrrr}
\hline
                     & MeCab     & Unigram   & BPE       & MaxMatch  & OpTok     & Random    \\ \hline
\multicolumn{7}{l}{\textit{\textbf{Accuracy of Machine (\%)}}}                                                 \\
BoW: Valid           & 42.48     & 43.97     & \b{44.15} & 43.61     & 43.37     & \u{41.94}     \\
BiLSTM: Valid        & 44.06     & \b{44.80}     & 44.53     & 44.74 & 44.30     & \u{42.39}     \\ \hline
\multicolumn{7}{l}{\textit{\textbf{Appropriateness (rank)}}}                                                   \\
Annotator1: Train    & \u{1.64}  & 2.74      & 3.23      & 3.02      & 2.54      & \b{5.59}      \\
Annotator1: Valid    & 3.63      & \u{1.94}  & 2.37      & 2.39      & 2.68      & \b{5.69}      \\
Annotator1: Total    & \u{1.86}  & 2.65      & 3.14      & 2.95      & 2.55      & \b{5.61}      \\\hdashline
Annotator2: Valid    & \u{2.34}  & 2.53      & 2.95      & 2.74      & 2.61      & \b{5.53}      \\\hdashline
Annotator3: Valid    & \u{2.41}  & 2.59      & 2.88      & 2.64      & 2.67      & \b{5.50}      \\ \hline
\multicolumn{7}{l}{\textit{\textbf{Accuracy of Human (\%)}}}                                                   \\
Annotator1: Train    & \b{93.90} & 93.85     & 93.89     & 93.68     & 93.69     & \u{93.23}     \\
Annotator1: Valid    & \b{97.50} & 97.41     & 97.32     & 97.50 & \u{97.05} & 97.23        \\
Annotator1: Total    & \b{94.30} & 94.25     & 94.27     & 94.10     & 94.06     & \u{93.67}     \\\hdashline
Annotator2: Train    & 96.08     & 96.04     & 95.86     & \b{96.14} & 95.97     & \u{95.65}     \\
Annotator2: Valid    & \b{93.66} & 92.94     & 92.94     & 93.03     & 93.12     & \u{92.94}     \\
Annotator2: Total    & \b{95.81} & 95.69     & 95.54     & 95.79     & 95.66     & \u{95.35}     \\ \hline
\multicolumn{7}{l}{\textit{\textbf{Response Time (sec. for all / correct answers)}}}                             \\
Annotator1: Train    & 6.15 / 5.65 & 6.06 / 5.56 & 6.13 / 5.63 & 6.12 / 5.61 & \u{6.05} / \u{5.53} & \b{6.99} / \b{6.36} \\
Annotator1: Valid    & 5.47 / 5.29 & \u{5.44} / 5.25 & 5.50 / 5.30 & 5.59 / 5.39 & 5.48 / \u{5.24} & \b{6.47} / \b{6.28} \\
Annotator1: Total    & 6.07 / 5.61 & \u{5.99} / 5.52 & 6.06 / 5.59 & 6.06 / 5.58 & 5.99 / \u{5.50} & \b{6.93} / \b{6.35} \\\hdashline
Annotator2: Train    & 5.08 / 4.75 & \u{5.06} / \u{4.75} & 5.09 / 4.77 & 5.09 / 4.79 & 5.11 / 4.78 & \b{5.68} / \b{5.33} \\
Annotator2: Valid    & 6.08 / \u{5.29} & \u{6.03} / 5.36 & 6.11 / 5.37 & 6.04 / 5.36 & 6.05 / 5.46 & \b{6.53} / \b{5.87} \\
Annotator2: Total    & 5.19 / 4.84 & \u{5.17} / \u{4.82} & 5.20 / 4.84 & 5.20 / 4.86 & 5.21 / 4.86 & \b{5.78} / \b{5.39} \\ \hline
\end{tabular}
\caption{
    Performance of machine learning models and annotation results for the JCommonsenseQA dataset with six tokenizers.
    The highest and lowest values in each raw are highlighted with bold and underline, respectively.
    \label{tbl:result}
}
\end{table*}

%% file: latex/5_conclusion.tex
\section{Conclusion}
\label{sec:conclusion}
This study examines preferred tokenization for human and ML models.
The tokenization preference was determined using the QA dataset and tokenizing the question texts with the six tokenization.
The annotation result quantitatively shows that preferred tokenization for ML models is not necessarily the same as the one for humans. 
Our contribution is to provide a quantitative examination of the intuition of the difference in preferred tokenization between humans and ML models.

This study still faces some unaddressed challenges.
For example, the annotation schema can be improved to increase the difference in human accuracy because we did not find considerable differences among tokenizations.
Some possible ways are to collect the answers by showing shuffled tokens of question texts in a flash card manner or by using eye-tracking systems.
This study is the first step to conduct such the advanced investigation.
